\documentclass[10pt,twocolumn,letterpaper]{article}

\usepackage{cvpr}
\usepackage{times}
\usepackage{epsfig}
\usepackage{graphicx}
\usepackage{amsmath}
\usepackage{amssymb}
\usepackage{enumitem}
\usepackage[normalem]{ulem}
\usepackage[ruled]{algorithm2e}
\usepackage{multirow}
\usepackage{array}
\usepackage{ctable} % for \specialrule command
\usepackage{makecell}
\usepackage[capposition=bottom]{floatrow}
\usepackage{comment}
\DeclareMathAlphabet{\mathcal}{OMS}{cmsy}{m}{n}
\usepackage{textcomp}
\usepackage{dblfloatfix}
\usepackage{caption}
\usepackage{subcaption}
\usepackage{colortbl,xcolor}
\usepackage{booktabs}
\usepackage{mathtools}
\usepackage{textpos}

\definecolor{customgray}{rgb}{0.9, 0.9, 0.9}
\newcolumntype{g}{>{\columncolor{customgray}}c}
\newcolumntype{z}{>{\columncolor{customgray}}l}
\newcolumntype{?}[1]{!{\vrule width #1}}
\renewcommand{\paragraph}[1]{{\vspace{2mm}\noindent\textbf{#1}\,\,}}

\usepackage[pagebackref=false,breaklinks=true,letterpaper=true,colorlinks,citecolor=blue,linkcolor=blue,bookmarks=false]{hyperref}
\cvprfinalcopy % *** Uncomment this line for the final submission

\def\cvprPaperID{2275} % *** Enter the CVPR Paper ID here

\ifcvprfinal\fi

\begin{document}

%%%%%%%%% TITLE
\title{Listen to Look: Action Recognition by Previewing Audio}
\author{Ruohan Gao\textsuperscript{1,2}\thanks{Work done during an internship at Facebook AI Research.}\hspace{10mm}Tae-Hyun Oh\textsuperscript{2}\thanks{T.-H. Oh is now with Dept. EE, POSTECH, Korea.}\hspace{10mm}Kristen Grauman\textsuperscript{1,2}
\hspace{10mm}Lorenzo Torresani\textsuperscript{2}\\
\textsuperscript{1}The University of Texas at Austin\hspace{10mm}\textsuperscript{2}Facebook AI Research\\
{\tt\small rhgao@cs.utexas.edu, \{taehyun,grauman,torresani\}@fb.com}
% For a paper whose authors are all at the same institution,
% omit the following lines up until the closing ``}''.
% Additional authors and addresses can be added with ``\and'',
% just like the second author.
% To save space, use either the email address or home page, not both
}

%\author{}

\maketitle
%\thispagestyle{empty}

%===========================================================
%Abstract
\begin{abstract}
In the face of the video data deluge, today's expensive clip-level classifiers are increasingly impractical. We propose a framework for efficient action recognition in untrimmed video that uses audio as a preview mechanism to eliminate both short-term and long-term visual redundancies. First, we devise an \textsc{ImgAud2Vid} framework that hallucinates clip-level features by distilling from lighter modalities---a single frame and its accompanying audio---reducing short-term temporal redundancy for efficient clip-level recognition. Second, building on \textsc{ImgAud2Vid}, we further propose \textsc{ImgAud-Skimming}, an attention-based long short-term memory network that iteratively selects useful moments in untrimmed videos, reducing long-term temporal redundancy for efficient video-level recognition. Extensive experiments on four action recognition datasets demonstrate that our method achieves the state-of-the-art in terms of both recognition accuracy and speed.
\end{abstract}
\begin{textblock*}{\textwidth}(0cm,-14.5cm)
\centering
In Proceedings of the IEEE Conference on Computer Vision and Pattern Recognition (CVPR), 2020.%
\end{textblock*}

%===========================================================
%Introduction
\section{Introduction}~\label{sec:intro}
With the growing popularity of portable image recording devices as well as online social platforms, internet users are generating and sharing an ever-increasing number of videos every day. According to a recent study, it would take a person over 5 million years to watch the amount of video that will be crossing global networks each month in 2021~\cite{cisco_study}. Therefore, it is imperative to devise systems that can recognize actions and events in these videos both accurately and efficiently. Potential benefits extend to many video applications, including video recommendation, summarization, editing, and browsing.

Recent advances in action recognition have mostly focused on building powerful clip-level models operating on short time windows of a few seconds~\cite{twostream,tran2015learning,feichtenhofer2016convolutional,carreira2017quo,wang2018non,feichtenhofer2019slowfast}. To recognize the action in a test video, most methods densely apply the clip classifier and aggregate the prediction scores of all the clips across the video. Despite encouraging progress, this approach becomes computationally impractical in real-world scenarios where the videos are untrimmed and span several minutes or even hours.

\begin{figure}
    \center
     \includegraphics[width=0.93\linewidth]{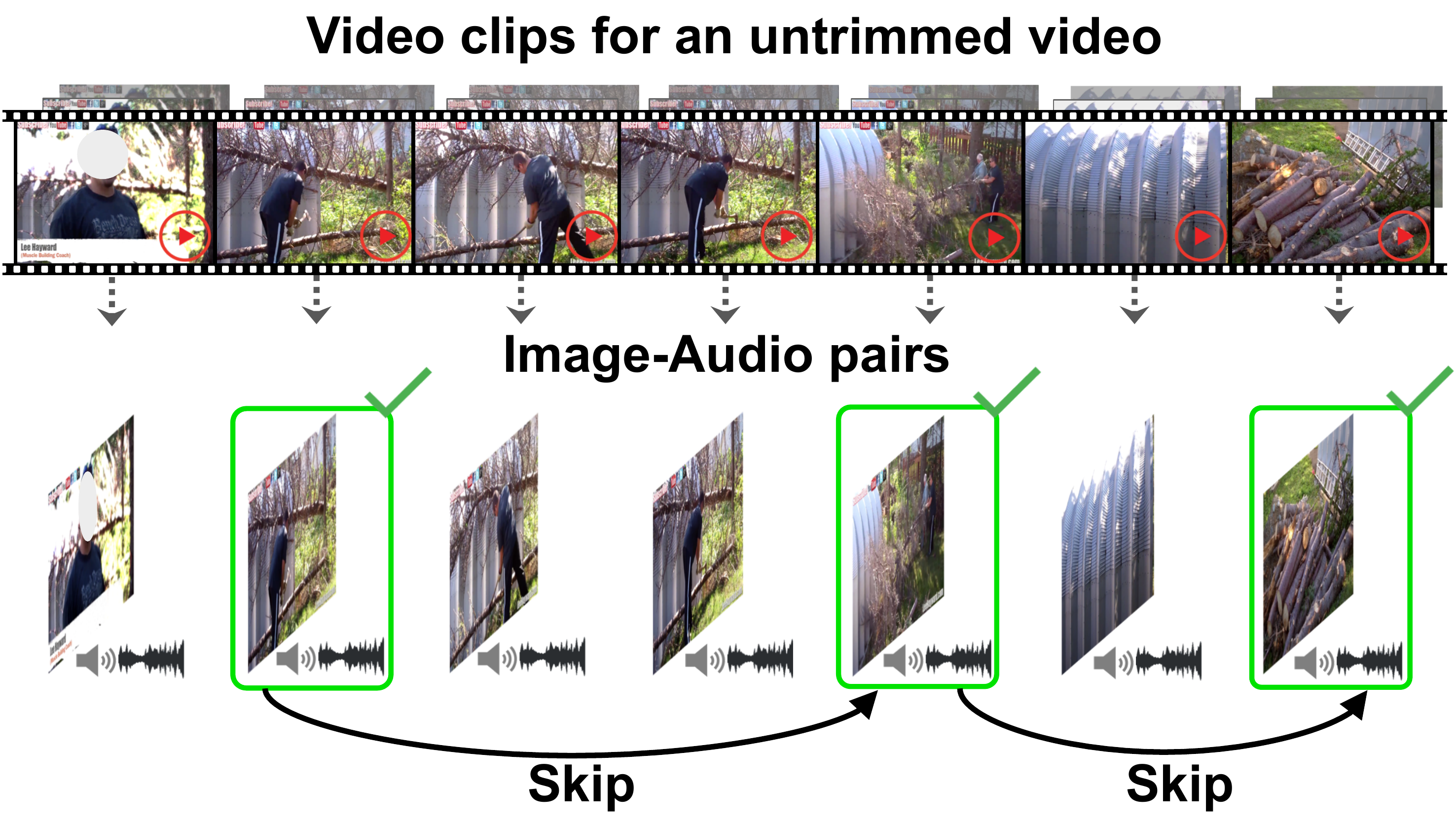}
    \caption{
    %Untrimmed video has high temporal redundancy within short-term clips as well as across the entire video. 
    Our approach learns to use audio as an efficient preview of the accompanying visual content, at two levels. First we replace the costly analysis of video clips with a more efficient processing of image-audio pairs. A single image captures most of the appearance information within the clip, while the audio provides important dynamic information. Then our video skimming module selects the key moments (a subset of image-audio pairs) to perform efficient video-level action recognition.
    \vspace{-0.2in}
    }
    \label{fig:concept_figure}
    \vspace{-0.1in}
\end{figure}

We contend that processing all frames or clips in a long untrimmed video may be unnecessary and even counter-productive. Our key insight is that there are two types of redundancy in video, manifested in both short-term clips as well as long-term periods. First, there is typically high temporal redundancy across the entire video (Fig.~\ref{fig:concept_figure}). Many clips capture the same event repetitively, suggesting it is unnecessary to process all the clips. Second, there is redundancy even within a clip: the visual composition within a short time span does not change abruptly; temporally adjacent frames are usually very similar, though there are temporal dynamics (motion) across frames. Therefore, it can be wasteful to process all clips and frames, especially when the video is very long. Moreover, for many activities, the actual actions taking place in the video can be very sparse. It is often a few important moments that are useful for recognition, while the rest actually distract the classifier. For example, in a typical video of surfing, a person might talk for a long time and prepare the equipment before he/she begins to surf.

Our idea is to use \emph{audio as an efficient video preview} to reduce both the clip-level and the video-level redundancy in long untrimmed videos. First, instead of processing a whole video clip, we propose an \textsc{ImgAud2Vid} teacher-student distillation framework to hallucinate a video descriptor (\eg, an expensive 3D CNN feature vector) from a single video frame and its accompanying audio. Based on our lightweight image-audio network, we further propose a novel attention-based long short-term memory (LSTM) network, called \textsc{ImgAud-Skimming}, which scans through the entire video and selects the key moments for the final video-level recognition. Both ideas leverage audio as a fast preview of the full video content. Our distilled image-audio model efficiently captures information over short extents, while the skimming module performs fast long-term modeling by skipping over irrelevant and/or uninformative segments across the entire video.

Audio has ideal properties to aid efficient recognition in long untrimmed videos: audio contains dynamics and rich contextual temporal information~\cite{gaver1993world} and, most importantly, it is much more computationally efficient to process compared to video frames. For example, as shown in Fig.~\ref{fig:concept_figure}, within a short clip of the action chopping wood, a single frame includes most of the appearance information contained in the clip, \ie, \{person, axe, tree\}, while the accompanying audio (the sound of the axe hitting the tree in this case) contains useful cues of temporal dynamics. Across the entire video, audio can also be beneficial to select the key moments that are useful for recognition. For example, the sound of the person talking initially can suggest that the actual action has not started, while the sound of the electric saw may indicate that the action is taking place. Our approach automatically learns such audio signals.

We experiment on four datasets (Kinetics-Sounds, Mini-Sports1M, ActivityNet, UCF-101) and demonstrate the advantages of our framework. Our main contributions are threefold. Firstly, we are the first to propose to replace the expensive extraction of clip descriptors with an efficient proxy distilled from audio. Secondly, we propose a novel video-skimming mechanism that leverages image-audio indexing features for efficient long-term modeling in untrimmed videos. Thirdly, our approach pushes the envelope of the trade-off between accuracy and speed favorably; we achieve state-of-the-art results on action recognition in untrimmed videos with few selected frames or clips.
%===========================================================
%Related Work
\section{Related Work}~\label{sec:related}
\vspace{-0.25in}

\paragraph{Action Recognition.} 
Action recognition in video has been extensively studied in the past decades. Research has transitioned from initial methods using hand-crafted local spatiotemporal features~\cite{laptev2003space,willems2008efficient,wang2013action} to mid-level descriptors~\cite{raptis2012discovering,jain2013representing,wang2013motionlets}, and more recently to deep video representations learned end-to-end~\cite{twostream,karpathy2014large,feichtenhofer2016convolutional}. Various deep networks have been proposed to model spatiotemporal information in videos~\cite{tran2015learning,carreira2017quo,qiu2017learning,wang2018non,feichtenhofer2019slowfast}.  Recent work includes capturing long-term temporal structure via recurrent  networks~\cite{yue2015beyond,donahue2015long} or ranking functions~\cite{fernando2015modeling}, pooling across space and/or time~\cite{wang2016temporal,Girdhar_17a_ActionVLAD}, modeling hierarchical or spatiotemporal information in videos~\cite{pirsiavash2014parsing,varol2016long}, building long-term temporal relations~\cite{wu2019long,zhou2018temporal}, or boosting accuracy by treating audio as another (late-fused) input modality~\cite{wu2016multi,long2018attention,wang2019makes,kazakos2019TBN}.

The above work focuses on building powerful models to improve recognition  without taking the computation cost into account, whereas our work aims to perform efficient action recognition in long untrimmed videos. Some work balances the accuracy-efficiency trade-off by using compressed video representations~\cite{wu2018compressed,shou2019dmc} or designing efficient network architectures~\cite{xie2018rethinking,zolfaghari2018eco,chen2018multi,tran2018closer,lin2019tsm}. In contrast, we propose to leverage audio to enable efficient clip-level and video-level action recognition in long untrimmed videos.

\vspace{-0.05in}
\paragraph{Action Proposal and Localization.} The goal of action localization~\cite{jain2014action,xu2017r,shou2017cdc,zhao2017temporal} is to find the temporal start and end of each action within a given untrimmed video and simultaneously recognize the action class. Many approaches~\cite{buch2017sst,xu2017r,lin2018bsn,wang2017untrimmednets} first use action proposals to identify candidate action segments. While reminiscent of our audio preview mechanism, the computational cost of most action proposal methods is several orders of magnitude larger. They generate a large number of clip proposals from the video, and then use flow~\cite{lin2018bsn} or deep features~\cite{buch2017sst,xu2017r}) for proposal selection. The selection stage is typically even more expensive than the final classification. Instead, our method addresses video classification, and high efficiency is a requirement in our design.

\vspace{-0.05in}
\paragraph{Audio-Visual Analysis.}
Recent work uses audio for an array of video understanding tasks outside of action recognition, including self-supervised representation learning~\cite{owens2016ambient,arandjelovic2017look,aytar2016soundnet,owens2018audio,Korbar2018cotraining,sun2019videobert}, audio-visual source separation~\cite{owens2018audio,afouras2018conversation,ephrat2018looking,gao2018objectSounds,zhao2018sound,gao2019coseparation}, localizing sounds in video frames~\cite{arandjelovic2017objects,Senocak_2019_PAMI,tian2018audio}, and generating sounds from video~\cite{owens2016visually,zhou2017visual,gao2019visualsound,morgadoNIPS18,Zhou_2019_ICCV}. Different from all the work above, we focus on leveraging audio 
for efficient action recognition. 

\vspace{-0.05in}
\paragraph{Cross-modal Distillation.}
Knowledge distillation~\cite{hinton2015distilling} addresses the problem of training smaller models from larger ones. We propose to distill the knowledge from an expensive clip-based model to a lightweight image-audio based model. Other forms of cross-modal distillation consider transferring supervision from RGB to flow or depth~\cite{gupta2016cross} or from a visual network to an audio network, or vice versa~\cite{aytar2016soundnet,owens2016ambient,albanie2018emotion,gan2019tracking}. In the opposite direction of ours, DistInit~\cite{girdhar2019distinit} performs uni-modal distillation from a pre-trained image model to a video model for representation learning from unlabeled video. Instead, we perform multi-modal distillation from a video model to an image-audio model for efficient clip-based action recognition.

\vspace{-0.05in}
\paragraph{Selection of Frames or Clips for Action Recognition.} 
Our approach is most related to the limited prior work on selecting salient frames or clips for action recognition in untrimmed videos. Whereas we use only weakly labeled video to train, some methods assume strong human annotations, \ie, ground truth temporal boundaries~\cite{yeung2016end} or sequential annotation traces~\cite{alwassel2018action}. Several recent methods~\cite{su2016leaving,fan2018watching,wu2019adaframe,wu2019multiagent} propose reinforcement learning (RL) approaches for video frame selection. Without using guidance from strong human supervision, they ease the learning process by restricting the agent to a rigid action space~\cite{fan2018watching}, guiding the selection process of the agent with a global memory module~\cite{wu2019adaframe}, or using multiple agents to collaboratively perform frame selection~\cite{wu2019multiagent}.

Unlike any of the above, we introduce a video skimming mechanism to select the key moments in videos aided by audio. We use audio as an efficient way to preview dynamic events for fast video-level recognition. Furthermore, our approach requires neither strong supervision nor complex RL policy gradients, which are often unwieldy to train. SCSampler~\cite{korbar2019scsampler} also leverages audio to accelerate action recognition in untrimmed videos. However, they only consider video-level redundancy by sampling acoustically or visually salient clips. In contrast, we address both clip-level and video-level redundancy, and we jointly learn the selection and recognition mechanisms. We include a comprehensive experimental comparison to methods in this genre. 

\vspace{-0.05in}
\paragraph{Video Summarization.} Video summarization work also aims to select keyframes or clips~\cite{lee2012discovering,gong2014diverse,mahasseni2017unsupervised,zhang2018retrospective}, but with the purpose of conveying the gist of the video to a human viewer. Instead, our work aims to select features useful for activity recognition. Beyond the difference in goal, our iterative attention-based mechanism is entirely novel as a frame selection technique.

%===========================================================
%Approach
\vspace{-0.05in}
\section{Approach}~\label{sec:approach}
Our goal is to perform accurate and efficient action recognition in long untrimmed videos. We first formally define our problem (Sec.~\ref{Sec:problem_formulation}); then we introduce how we use audio as a clip-level preview to hallucinate video descriptors based on only a single static frame and its accompanying audio segment (Sec.~\ref{Sec:clip_level_app}); finally we present how we leverage image-audio indexing features to obtain a video-level preview, and learn to skip over irrelevant or uninformative segments in untrimmed videos (Sec.~\ref{Sec:video_level_app}). 

\subsection{Problem Formulation}~\label{Sec:problem_formulation}
Given a long untrimmed video $\mathcal{V}$, the goal of video classification is to classify $\mathcal{V}$ into a predefined set of $C$ classes. Because $\mathcal{V}$ can be very long, it is often intractable to process all the video frames together through a single deep network due to memory constraints. Most current approaches~\cite{twostream,karpathy2014large,tran2015learning,carreira2017quo,qiu2017learning,tran2018closer,wang2018non,feichtenhofer2019slowfast} first train a clip-classifier $\Omega(\cdot)$ to operate on a short fixed-length video clip $\mathbf{V} \in \mathbb{R}^{F \times 3 \times H \times W}$ of $F$  frames with spatial resolution $H \times W$, typically spanning several seconds. Then, given a test video of arbitrary length, these methods densely apply the clip-classifier to $N$ clips $\{\mathbf{V}_1, \mathbf{V}_2, \ldots, \mathbf{V}_N\}$ which are taken at a fixed hop size across the entire video. The final video-level prediction is obtained by aggregating the clip-level predictions of all $N$ clips.

As discussed in Sec.~\ref{sec:intro}, such paradigms for video recognition are highly inefficient at two levels: (1) \emph{clip-level}---within each short clip $\mathbf{V}$, temporally close frames are visually similar, and (2) \emph{video-level}---across all the clips in $\mathcal{V}$, often only a few clips contain the key moments for recognizing the action. Our approach addresses both levels of redundancy via novel uses of audio.

Each video clip $\mathbf{V}$ is accompanied by an audio segment $\mathbf{A}$. The starting frame $\mathbf{I}$ among the $F$ frames within the short clip $\mathbf{V}$ usually contains most of the appearance cues already, while the audio segment $\mathbf{A}$ contains rich contextual temporal information (recall the wood cutting example in Fig.~\ref{fig:concept_figure}). Our idea is to replace the powerful but expensive clip-level classifier $\Omega(\cdot)$ that takes $F$ frames as input with an efficient image-audio classifier $\Phi(\cdot)$ that only takes the starting frame $\mathbf{I}$ and its accompanying audio segment $\mathbf{A}$ as input, while preserving the clip-level information as much as possible. Namely, we seek to learn $\Phi(\cdot)$ such that
\vspace{-0.05in}
\begin{equation}
	\Omega(\mathbf{V}_j) \approx \Phi(\mathbf{I}_j, \mathbf{A}_j),\quad j \in \{1,2, \ldots, N\},
\vspace{-0.05in}
\end{equation}
for a given pre-trained clip-classifier $\Omega(\cdot)$. In Sec.~\ref{Sec:clip_level_app}, we design an \textsc{ImgAud2Vid} distillation framework to achieve this goal. Through this step, we replace the processing of high-dimensional video clips $\{\mathbf{V}_1, \mathbf{V}_2, \ldots, \mathbf{V}_N\}$ with a lightweight model analyzing compact image-audio pairs $\{(\mathbf{I}_1, \mathbf{A}_1), (\mathbf{I}_2, \mathbf{A}_2), \ldots,(\mathbf{I}_N, \mathbf{A}_N)\}$. 

Next, building on our efficient image-audio classifier $\Phi(\cdot)$, to address video-level redundancy we design an attention-based LSTM network called \textsc{ImgAud-Skimming}. Instead of classifying every image-audio pair using $\Phi(\cdot)$ and aggregating all their prediction results, our \textsc{ImgAud-Skimming} framework iteratively selects the most useful image-audio pairs. Namely, our method efficiently selects a small subset of $T$ image-audio pairs from the entire set of $N$ pairs in the video (with $T \ll N$) and only aggregates the predictions from these selected pairs. We present our video skimming mechanism in Sec.~\ref{Sec:video_level_app}.

\subsection{Clip-Level Preview}~\label{Sec:clip_level_app}
\begin{figure}
    \center
    \includegraphics[width=0.95\linewidth]{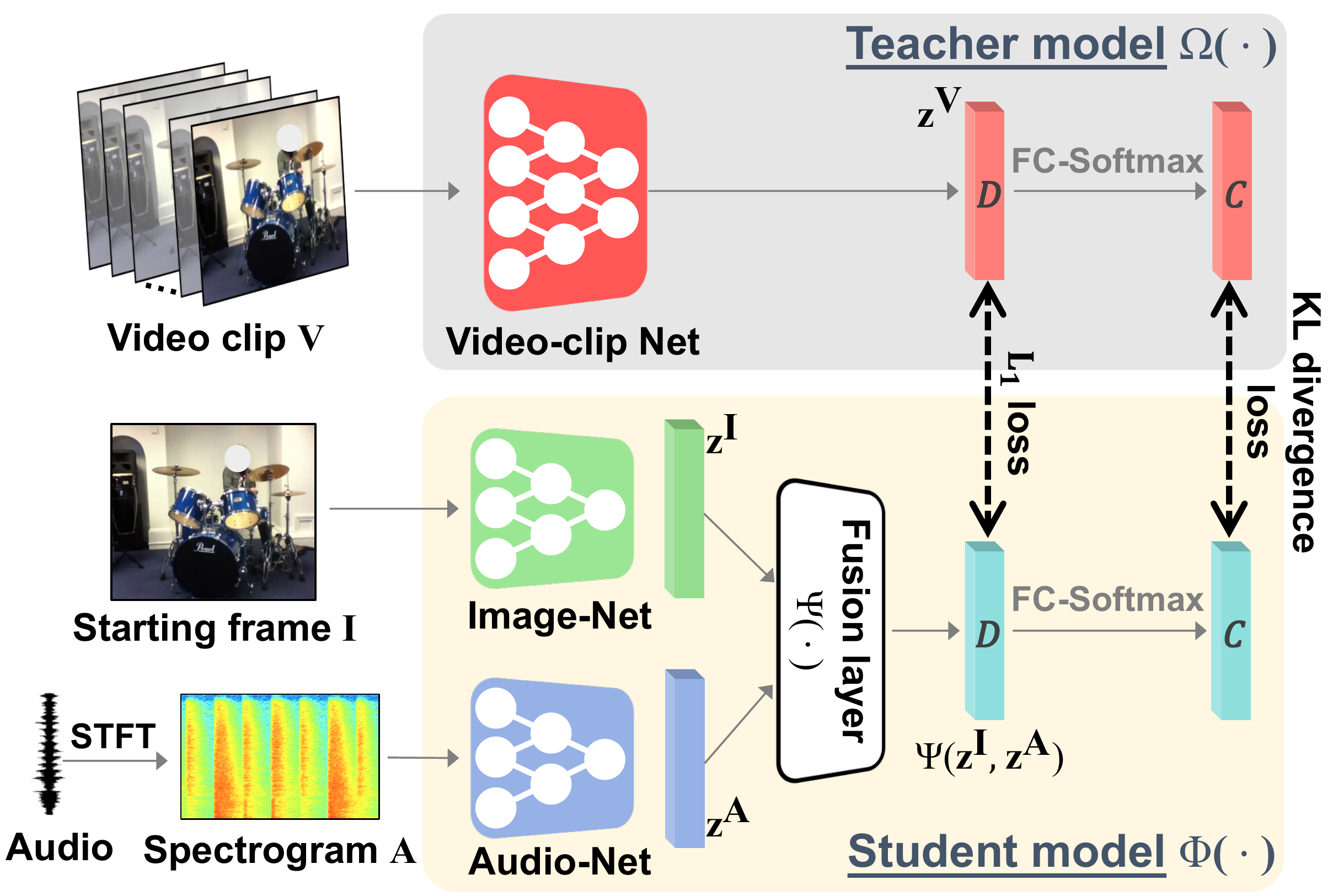}
    \caption{\textsc{ImgAud2Vid} distillation framework: The teacher model is a video-clip classifier, and the student model consists of a visual stream that takes the starting frame of the clip as input and an audio stream that takes the audio spectrogram as input. By processing only a single frame and the clip's audio, we get an estimate of what the expensive video descriptor would be for the full clip.
    \vspace{-0.1in}}
    \label{fig:imgAud2vid_network}
    \vspace{-0.1in}
\end{figure}
We present our approach to perform efficient clip-level recognition and our \textsc{ImgAud2Vid} distillation network architecture.
As shown in Fig.~\ref{fig:imgAud2vid_network}, the clip-based model takes a video clip $\mathbf{V}$ of $F$ frames as input and based on that video volume generates a clip descriptor $\mathbf{z}^\mathbf{V}$ of dimensionality $D$. A fully-connected layer is used to make predictions among the $C$ classes in Kinetics. For the student model, we use a two-stream network: the image stream takes the first frame $\mathbf{I}$ of the clip as input and extracts an image descriptor $\mathbf{z}^\mathbf{I}$; the audio stream takes the audio spectrogram $\mathbf{A}$ as input and extracts an audio feature vector $\mathbf{z}^\mathbf{A}$. We concatenate $\mathbf{z}^\mathbf{I}$~and~$\mathbf{z}^\mathbf{A}$ to generate an image-audio feature vector of dimensionality $D$ using a fusion network $\Psi(\cdot)$ that consists of two fully-connected layers. A final fully-connected layer is used to produce a $C$-class prediction like the teacher model.

The teacher model $\Omega(\cdot)$ returns a softmax distribution over $C$ classification labels. These predictions are used as soft targets for training weights associated with the student network $\Phi(\cdot)$ using the following objective:
\vspace{-0.05in}
\begin{equation}
	\mathcal{L}_{\text{KL}} = - \sum\nolimits_{\{(\mathbf{V},\mathbf{I},\mathbf{A})\}}\sum\nolimits_{c} \Omega_c(\mathbf{V}) \log \Phi_c (\mathbf{I}, \mathbf{A}),
	\vspace{-0.05in}
\end{equation}
where $\Omega_c(\mathbf{V})$ and $\Phi_c (\mathbf{I}, \mathbf{A})$ are the softmax scores of class $c$ for the teacher model and the student model, respectively. We further impose an $\mathcal{L}_1$ loss on the clip descriptor $\mathbf{z}^\mathbf{V}$ and the image-audio feature to regularize the learning process:
\vspace{-0.05in}
\begin{equation}
	\mathcal{L}_1 = \sum\nolimits_{\{(\mathbf{z}^\mathbf{V},\mathbf{z}^\mathbf{I},\mathbf{z}^\mathbf{A})\}} \|\mathbf{z}^\mathbf{V}  - \Psi(\mathbf{z}^\mathbf{I}, \mathbf{z}^\mathbf{A})\|_1.
	\vspace{-0.05in}
\end{equation}
The final learning objective for \textsc{ImgAud2Vid} distillation is a combination of these two losses:
\vspace{-0.05in}
\begin{equation}~\label{equation:distillation}
	\mathcal{L}_{\text{Dist.}} = \mathcal{L}_1 +  \lambda L_\text{{KL}},
\vspace{-0.05in}
\end{equation}
where $\lambda$ is the weight for the KL divergence loss. The training is done over the image and audio student networks (producing representations $\mathbf{z}^\mathbf{I}$ and $\mathbf{z}^\mathbf{A}$, respectively) and the fusion model $\Psi(\cdot)$ with respect to a fixed teacher video-clip model. The teacher model we use is a R(2+1)D-18~\cite{tran2018closer} video-clip classifier, which is pre-trained on Kinetics~\cite{kay2017kinetics}. Critically, processing the audio for a clip is substantially faster than processing all its frames, making audio an efficient preview. See Sec.~\ref{Sec:clip_level_results} for cost comparisons. After distillation, we fine-tune the student model on the target dataset to perform efficient clip-level action recognition. 

\subsection{Video-Level Preview}~\label{Sec:video_level_app}
\textsc{ImgAud2Vid} distills knowledge from a powerful clip-based model to an efficient image-audio based model. Next, we introduce how we leverage the distilled image-audio network to perform efficient video-level recognition. Recall that for long untrimmed video, processing only a subset of clips is desirable both for speed and accuracy, \ie, to ignore irrelevant content.

\begin{figure*}
    \center
    \includegraphics[width=0.97\linewidth]{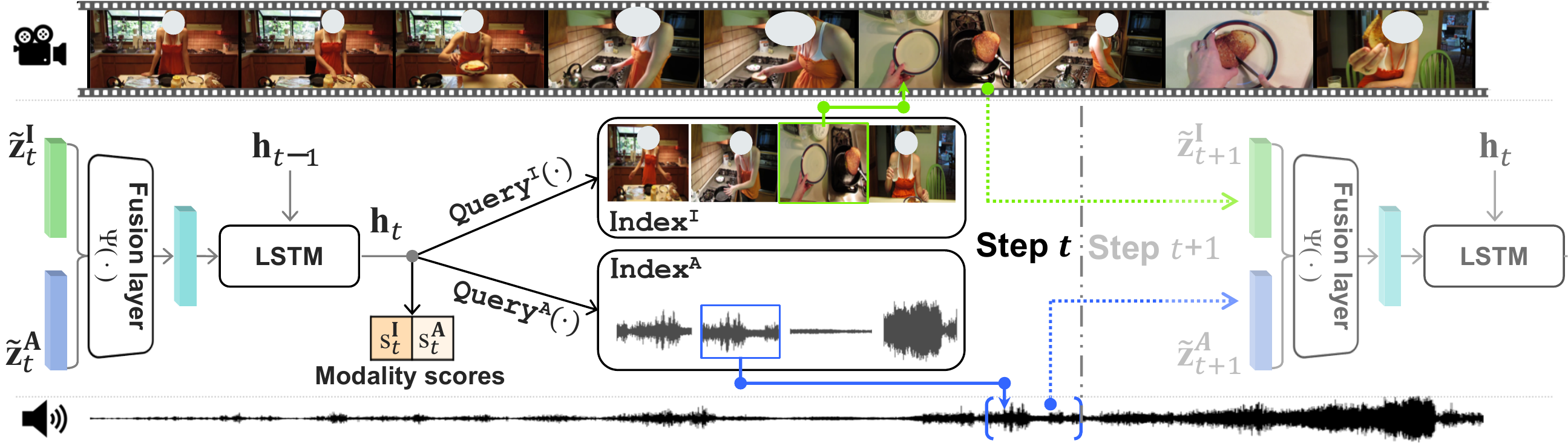}
    \caption{Our \textsc{ImgAud-Skimming} network is an LSTM network that interacts with the sequences of image and audio indexing features to select where to ``look at" and ``listen to" next. At each time step, it takes the image feature and audio feature for the current time step as well as the previous hidden state and cell output as input, and produces the current hidden state and cell output. The hidden state for the current time step is used to make predictions about the next moment to focus on in the untrimmed video through the querying operation illustrated in Fig.~\ref{fig:querying_mechanism}. The average-pooled \textsc{ImgAud2Vid} features of all selected time steps is used to make the final prediction of the action in the video.
    \vspace{-0.1in}
    }
    \label{fig:skimming_network}
    \vspace{-0.1in}
\end{figure*}

We design \textsc{ImgAud-Skimming}, an attention-based LSTM network (Fig.~\ref{fig:skimming_network}), which interacts with the sequence of image-audio pairs $\{(\mathbf{I}_1, \mathbf{A}_1), (\mathbf{I}_2, \mathbf{A}_2), \ldots, (\mathbf{I}_N, \mathbf{A}_N)\}$, whose features are denoted as $\{\mathbf{z}^\mathbf{I}_1, \mathbf{z}^\mathbf{I}_2, \ldots, \mathbf{z}^\mathbf{I}_N \}$ and $\{\mathbf{z}^\mathbf{A}_1, \mathbf{z}^\mathbf{A}_2, \ldots, \mathbf{z}^\mathbf{A}_N \}$, respectively. At the $t$-th time step, the LSTM cell takes the \emph{indexed} image feature $\mathbf{\tilde z}^\mathbf{I}_t$ and the \emph{indexed} audio feature $\mathbf{\tilde z}^\mathbf{A}_t$, as well as the previous hidden state $\textbf{h}_{t-1}$ and the previous cell output $\textbf{c}_{t-1}$ as input, and produces the current hidden state $\textbf{h}_t$ and the cell output $\textbf{c}_t$:
\vspace{-0.05in}
\begin{equation}
	\textbf{h}_t, \textbf{c}_t = \mathtt{LSTM}\big(\Psi(\mathbf{\tilde z}^\mathbf{I}_t, \mathbf{\tilde z}^\mathbf{A}_t), \textbf{h}_{t-1}, \textbf{c}_{t-1}\big),
\vspace{-0.05in}
\end{equation}
where $\Psi(\cdot)$ is the same fusion network used in \textsc{ImgAud2Vid} with the same parameters. To fetch the indexed features $\mathbf{\tilde z}^\mathbf{I}_t$ and $\mathbf{\tilde z}^\mathbf{A}_t$ from the feature sequences, an indexing operation is required. This operation is typically non-differentiable. Instead of relying on approximating policy gradients as in prior work~\cite{fan2018watching,wu2019adaframe,wu2019multiagent}, we propose to deploy a differentiable soft indexing mechanism, detailed below.

We predict an image query vector $\mathbf{q}^{\mathbf{I}}_{t}$ and an audio query vector $\mathbf{q}^{\mathbf{A}}_{t}$ from the hidden state $\mathbf{h}_t$ at each time step through two prediction networks $\mathtt{Query}^{\mathbf{I}}(\cdot)$ and $\mathtt{Query}^{\mathbf{A}}(\cdot)$. The query vectors, $\mathbf{q}^{\mathbf{I}}_{t}$ and $\mathbf{q}^{\mathbf{A}}_{t}$, are used to query the respective sequences of image indexing features $\{\mathbf{z}^\mathbf{I}_1, \mathbf{z}^\mathbf{I}_2, \ldots, \mathbf{z}^\mathbf{I}_N \}$ and audio indexing features $\{\mathbf{z}^\mathbf{A}_1, \mathbf{z}^\mathbf{A}_2, \ldots, \mathbf{z}^\mathbf{A}_N \}$. The querying operation is intended to predict which part of the untrimmed video is more useful for recognition of the action in place and decide where to ``look at" and ``listen to" next. It is motivated by attention mechanisms~\cite{graves2014neural,sukhbaatar2015end,vinyals2015pointer,vaswani2017attention}, but we adapt this scheme to the problem of selecting useful moments for action recognition in untrimmed video. 

\begin{figure}
    \center
    \includegraphics[width=\linewidth]{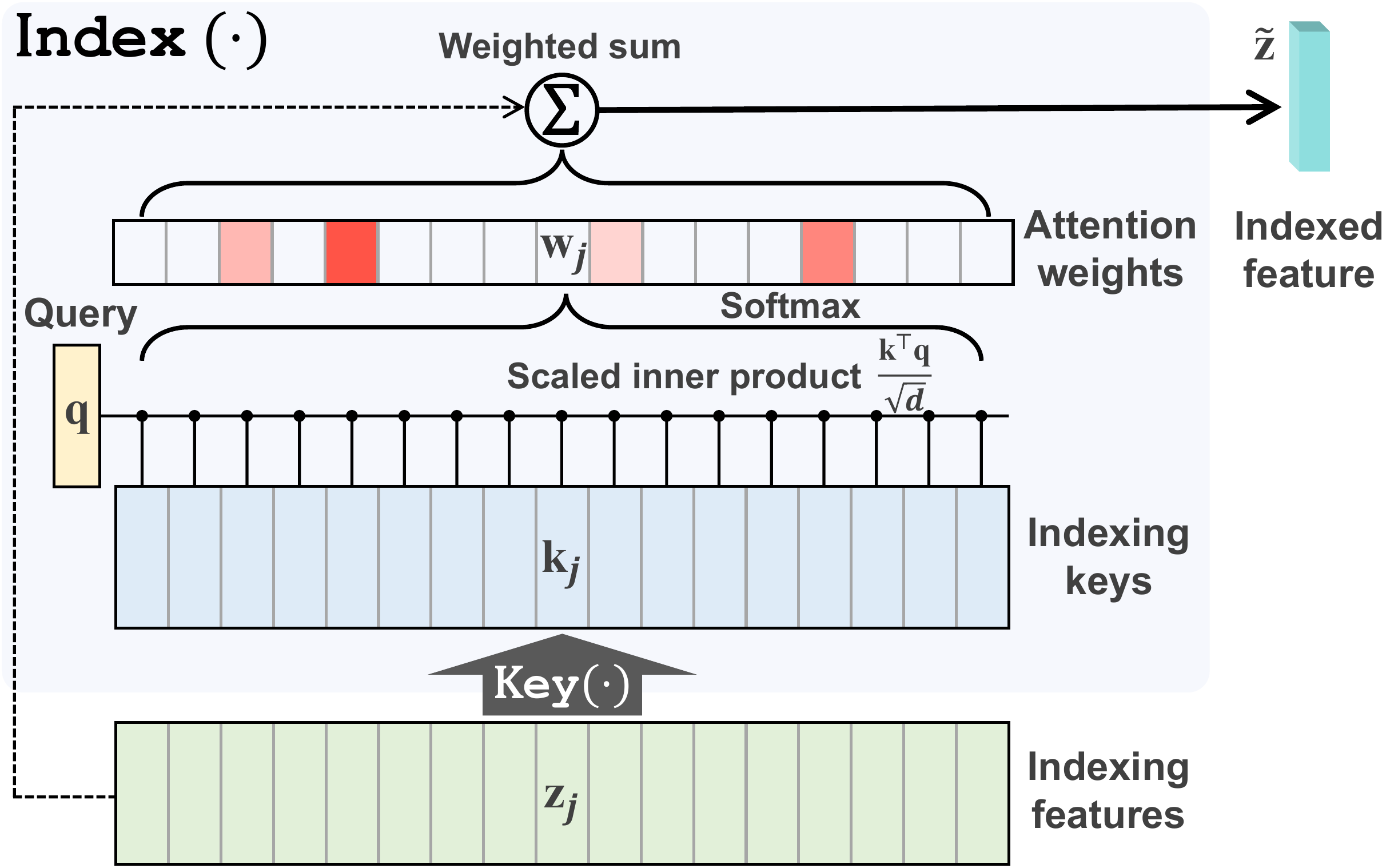}
    \caption{Attention-based frame selection mechanism.
    \vspace{-0.15in}
    }
    \label{fig:querying_mechanism}
    \vspace{-0.1in}
\end{figure}

Figure~\ref{fig:querying_mechanism} illustrates our querying mechanism. First, we use one fully-connected layer $\mathtt{Key}(\cdot)$ to transform indexing features $\mathbf{z}$ to indexing keys $\mathbf{k}$. Then, we get an attention score $\frac{\mathbf{k}^\top\mathbf{q}}{\sqrt{d}}$ for each indexing key in the sequence, where $d$ is the dimensionality of the key vector. A $\mathtt{Softmax}$ layer normalizes the attention scores and generates an attention weight vector $\mathbf{w}$ by:
\begin{equation}
\mathbf{w} =  \mathtt{Softmax}\Big( \frac{[\mathbf{k}_1 \mathbf{k}_2 \ldots \mathbf{k}_N]^\top \cdot \mathbf{q}}{\sqrt{d}} \Big),
\end{equation}
where $\mathbf{k}_j = \mathtt{Key}(\mathbf{z}_j)$, $j \in \{1,2, \ldots, N\}$.

At each time step $t$ (we omit $t$ for simplicity if deducible), one could obtain the frame index for the next time step by $\arg\max (\mathbf{w})$. However, this operation is not differentiable. Instead of directly using the image and audio features of the selected frame index, we use the weighted average of the sequence of indexing features to generate an aggregated feature vector 
$\mathbf{\tilde z}^\mathbf{I}_{t+1} = \mathtt{Index}^\mathbf{I}(\mathbf{w}_{t})$ and
$\mathbf{\tilde z}^\mathbf{A}_{t+1} = \mathtt{Index}^\mathbf{A}(\mathbf{w}_{t})$
as input to the fusion network $\Psi(\cdot)$, where
\begin{equation}
\begin{aligned}
&\mathtt{Index}^\mathbf{I}(\mathbf{w}) \coloneqq \textstyle\sum_{j=1}^{N} w_j \mathbf{z}^\mathbf{I}_j, 
&\\
&\mathtt{Index}^\mathbf{A}(\mathbf{w}) \coloneqq \textstyle\sum_{j=1}^{N} w_j \mathbf{z}^\mathbf{A}_j, & w_{j\in\{1,\cdots,N\}}\in\mathbb{R}_+.
\end{aligned}
\end{equation}

The querying operations are performed independently on the visual and audio modalities, and produce distinct weight vectors $\mathbf{w}^{\mathbf{I}}_t$ and $\mathbf{w}^{\mathbf{A}}_t$ to find the visually-useful and acoustically-useful moments, respectively. These two weight vectors may give importance to different moments in the sequence. We fuse this information by dynamically adjusting how much to rely on each modality at each step. 
To this end, we predict two modality scores $s^{\mathbf{I}}_t$ and $s^{\mathbf{A}}_t$, from the hidden state $\mathbf{h}_t$ through a two-way classification layer. $s^\mathbf{I}_t$ and $s^\mathbf{A}_t$ ($s^\mathbf{I}_t, s^\mathbf{A}_t \in [0,1], s^\mathbf{I}_t + s^\mathbf{A}_t = 1$) indicate how much the system decides to rely on the visual modality versus the audio modality, respectively, at time step $t$. Then, the image and audio feature vectors for the next time step are finally obtained by aggregating the feature vectors predicted both visually and acoustically, as follows:
\begin{equation}
\begin{aligned}
\mathbf{\tilde z}^\mathbf{I}_{t+1} = s^{\mathbf{I}}_t \cdot \mathtt{Index}^\mathbf{I}(\mathbf{w}^{\mathbf{I}}_{t}) + s^{\mathbf{A}}_t \cdot \mathtt{Index}^\mathbf{I}(\mathbf{w}^{\mathbf{A}}_t),
&\\
\mathbf{\tilde z}^\mathbf{A}_{t+1} = s^{\mathbf{I}}_t \cdot \mathtt{Index}^\mathbf{A}(\mathbf{w}^{\mathbf{I}}_{t}) + s^{\mathbf{A}}_t \cdot \mathtt{Index}^\mathbf{A}(\mathbf{w}^{\mathbf{A}}_t).
\end{aligned}
\end{equation}

Motivated by iterative attention~\cite{mnih2014recurrent}, we repeat the above procedure for $T$ steps, and average the image-audio features obtained. Namely,
\begin{equation}~\label{equation:feature_pooling}
	\mathbf{m} =  \tfrac{1}{T} {\textstyle \sum_{j=1}^{T}} \Psi (\mathbf{\tilde z}^\mathbf{I}_j, \mathbf{\tilde z}^\mathbf{A}_j).
\end{equation}
$\mathbf{m}$ is a feature summary of the useful moments selected by \textsc{ImgAud-Skimming}. A final fully-connected layer followed by $\mathtt{Softmax}(\cdot)$ takes $\mathbf{m}$ as input and makes predictions of action categories. The network is then trained with cross-entropy loss and video-level action label annotations.

While we optimize the \textsc{ImgAud-Skimming} network for a fixed number of $T$ steps during training, at inference time we can stop early at any step depending on the computation budget. Moreover, instead of using all indexing features, we can also use a subset of indexing features to accelerate inference with the help of feature interpolation. See Sec.~\ref{Sec:video_level_results} for details about the efficiency and accuracy trade-off when using sparse indexing features and early stopping. 
%===========================================================
%Experiments
\section{Experiments}\label{sec:results}
\vspace{-0.05in}

Using a total of 4 datasets, we evaluate our approach for accurate and efficient clip-level action recognition (Sec.~\ref{Sec:clip_level_results}) and video-level action recognition (Sec.~\ref{Sec:video_level_results}).

\vspace{-0.05in}
\paragraph{Datasets:} Our distillation network is trained on Kinetics~\cite{kay2017kinetics}, and we evaluate on four other datasets: Kinetics-Sounds~\cite{arandjelovic2017look}, UCF-101~\cite{soomro2012ucf101}, ActivityNet~\cite{caba2015activitynet}, and Mini-Sports1M~\cite{karpathy2014large}. Kinetics-Sounds and UCF-101 contain only short trimmed videos, so we only test on them for clip-level recognition; ActivityNet contains videos of various lengths, so it is used as our main testbed for both clip-level and video-level recognition; Mini-Sports1M contains only long untrimmed videos, and we use it for evaluation of video-level recognition. See Supp. for details of these datasets.

\vspace{-0.05in}
\paragraph{Implementation Details:} We implement in PyTorch. For \textsc{ImgAud2Vid}, the R(2+1)D-18~\cite{tran2018closer} teacher model takes 16 frames of size $112 \times 112$ as input. The student model uses a ResNet-18 network for both the visual and audio streams, which take the starting RGB frame of size $112 \times 112$ and a 1-channel audio-spectrogram of size $101 \times 40$ (1 sec.~audio segment) as input, respectively. We use $\lambda=100$ for the distillation loss in Equation~\ref{equation:distillation}. For \textsc{ImgAud-Skimming}, we use a one-layer LSTM with 1,024 hidden units and a dimension of 512 for the indexing key vector.  We use $T = 10$ time steps during training. See Supp. for details.

\vspace{-0.05in}
\subsection{Clip-level Action Recognition}~\label{Sec:clip_level_results}
\vspace{-0.2in}

First, we directly evaluate the performance of the image-audio network distilled from the video model. We fine-tune on each of the three datasets for clip-level recognition and compare against the following baselines:

\vspace{-0.1in}
\begin{itemize}[leftmargin=4mm]
\itemsep0em
\item \textbf{Clip-based Model:} The R(2+1)D-18 teacher model.
\vspace{-0.05in}
\item \textbf{Image-based Model (distilled/undistilled):} A ResNet-18 frame-based model. The undistilled model is pre-trained on ImageNet, and the distilled model is similar to our method except that the distillation is performed using only the visual stream. 
\vspace{-0.05in}
\item \textbf{Audio-based Model (distilled/undistilled):} The same as the image-based model except here we only use the audio stream for recognition and distillation. The model is pre-trained on ImageNet to accelerate convergence.
\vspace{-0.05in}
\item \textbf{Image-Audio Model (undistilled):} The same image-audio network as our method but without distillation.
\end{itemize}
\vspace{-0.1in}

\begin{figure}
    \center
    \includegraphics[scale=0.53]{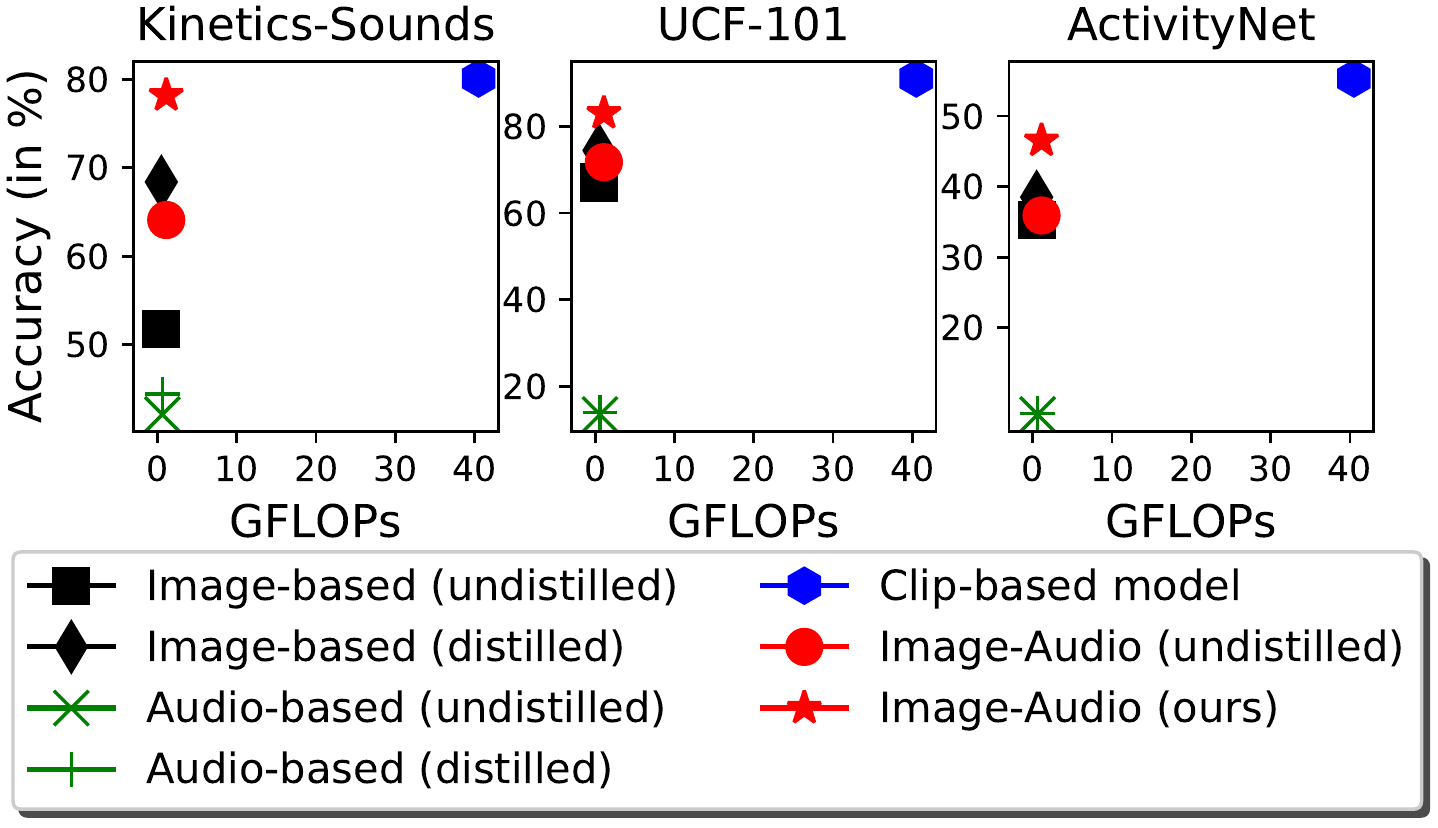}
    \caption{Clip-level action recognition on Kinetics-Sounds, UCF-101, and ActivityNet. We compare the recognition accuracy and the computational cost of our model against a series of baselines. Our \textsc{ImgAud2Vid} approach strikes a favorable balance between accuracy and efficiency.
    \vspace{-0.1in}} 
    \label{fig:clip_level_comparison}
    \vspace{-0.1in}
\end{figure}

For each baseline, we use the corresponding model as initialization and fine-tune on the same target dataset for clip-based action recognition. Note that our purpose here is not to compete on recognition accuracy using R(2+1)D-18 (or any other expensive video features), but rather to demonstrate our distilled image-audio features can approximate its recognition accuracy much more efficiently.

Figure~\ref{fig:clip_level_comparison} compares the accuracy vs.~efficiency for our approach and the baselines. Our distilled image-audio network achieves accuracy comparable to that of the clip-based teacher model, but at a much reduced computational cost. Moreover, the models based on image-only or audio-only distillation produce lower accuracy. This shows that the image or audio alone is not sufficient to hallucinate the video descriptor, but when combined they provide sufficiently complementary information to reduce the accuracy gap with the true (expensive) video-clip descriptor. 

To understand when audio helps the most, we compute the $\mathcal{L}_1$ distance of the hallucinated video descriptor to the ground-truth video descriptor by our \textsc{ImgAud2Vid} distillation and the image-based distillation. The top clips for which we best match the ground-truth tend to be dynamic scenes that have informative audio information, \eg, grinding meat, jumpstyle dancing, playing cymbals, playing bagpipes, wrestling, and welding. See Supp. for examples.

\begin{table*}
\resizebox{1\linewidth}{!}{
\begin{tabular}{@{}r|cgcgcgcgcgc@{}}
\toprule
& \textsc{Random} & \textsc{Uniform} & \textsc{Front} & \textsc{Center} & \textsc{End} & \textsc{SCSampler}~\cite{korbar2019scsampler} & \textsc{Dense} & \textsc{LSTM} &  \textsc{Non-Recurrent} & Ours (sparse / dense) \\
\hline
ActivityNet &  63.7     &     64.8     &    39.0   &    59.0  &            38.1         &     69.1      &    66.3  & 63.5 & 67.5  & \textbf{70.3} / \textbf{71.1} \\ 
Mini-Sports1M       &  35.4      &  35.6       &   17.1         &  29.7   &  17.4  &     38.4       & 37.3    &  34.1          & 38.0  &  \textbf{39.2} / \textbf{39.9} \\ \bottomrule
\end{tabular}
}
\vspace{-1mm}
\caption{Video-level action recognition accuracy (in \%) on ActivityNet (\# classes: 200) and Mini-Sports1M (\# classes: 487). Kinetics-Sounds and UCF-101 consist of only short trimmed videos, so they are not applicable here. Our method consistently outperforms all baseline methods. Ours (sparse) uses only about 1/5 the computation cost of the last four baselines, while achieving large accuracy gains. See Table~\ref{Table:activitynet_sota} for more computation cost comparisons.
\vspace{-0.1in}
}
\vspace{-0.05in}
\label{Table:video_level_comparison}
\end{table*}

\vspace{-0.05in}
\subsection{Untrimmed Video Action Recognition}~\label{Sec:video_level_results}
\vspace{-0.2in}

Having demonstrated the clip-level performance of our distilled image-audio network, we now examine the impact of the $\textsc{ImgAud-Skimming}$ module on video-level recognition. We evaluate on ActivityNet~\cite{caba2015activitynet} and Mini-Sports1M~\cite{karpathy2014large}, which contain long untrimmed videos. 

\begin{figure}
  \begin{subfigure}[b]{0.5\textwidth}
    \includegraphics[width=\textwidth]{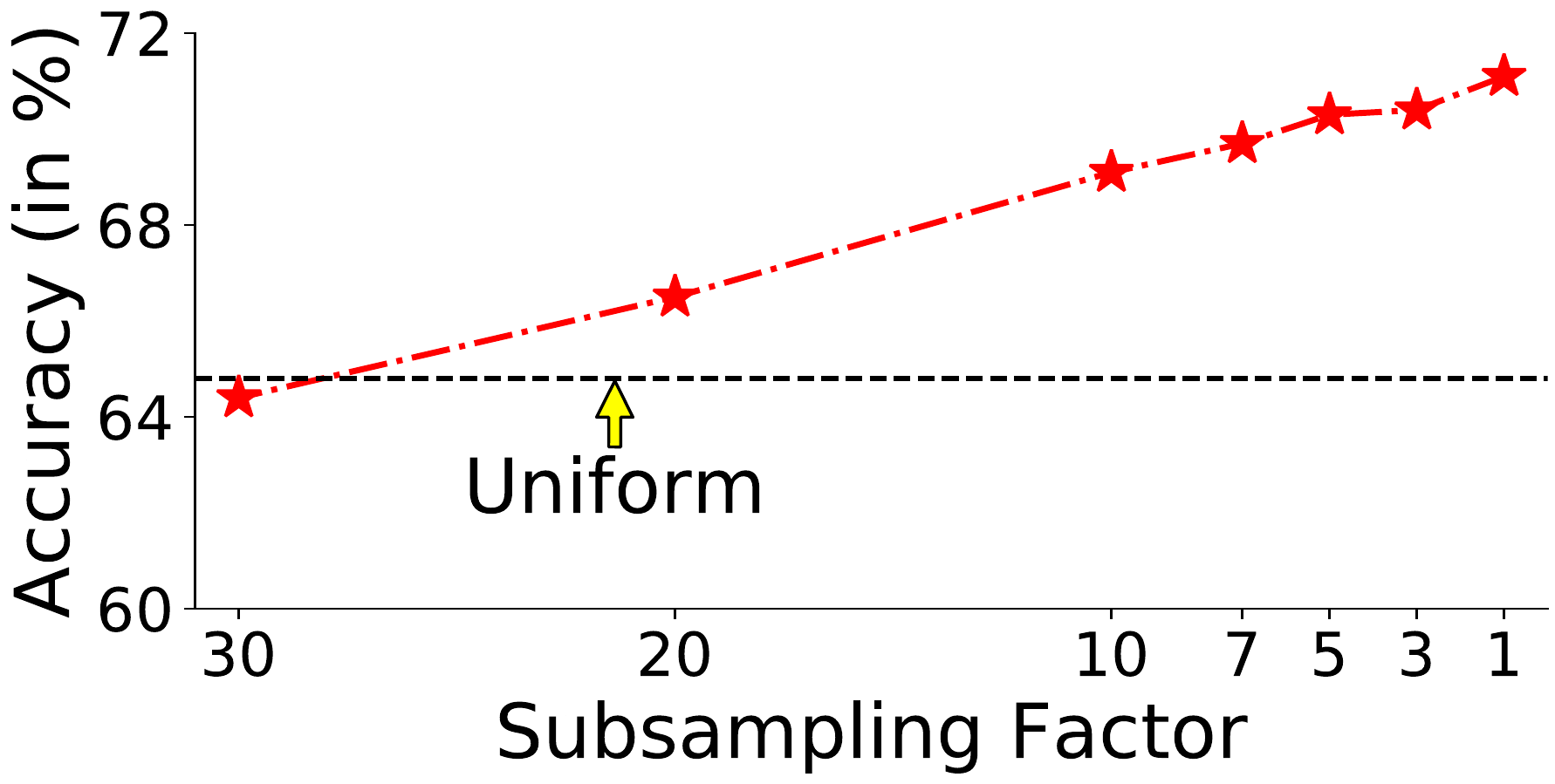}
     \vspace{-0.2in}
    \caption{Feature interpolation}
    \label{fig:interpolation}
  \end{subfigure}
  \begin{subfigure}[b]{0.49\textwidth}
    \includegraphics[width=\textwidth]{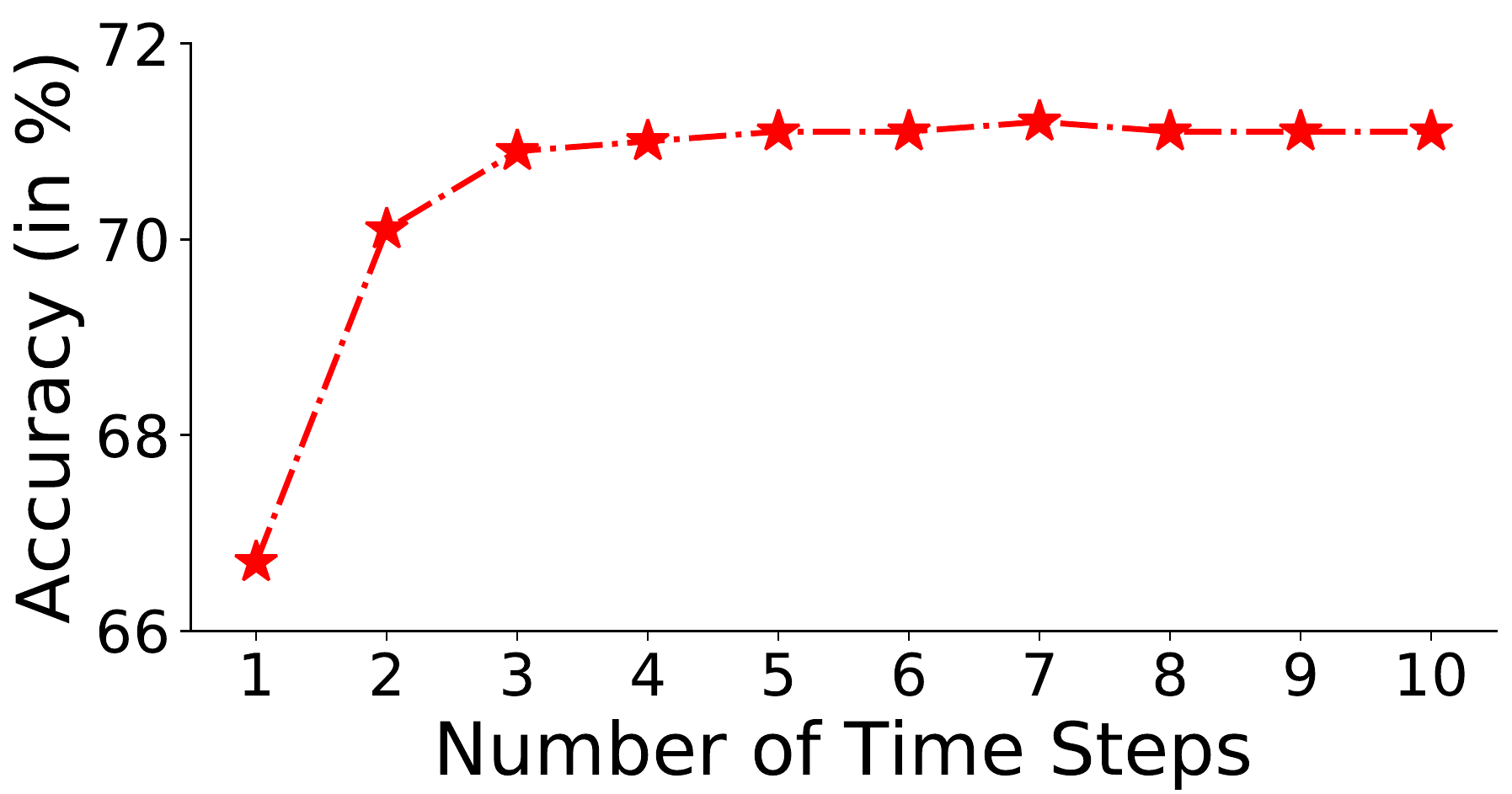}
        \vspace{-0.2in}
    \caption{Early stopping}
    \label{fig:early_stop}
  \end{subfigure}
  \caption{Trade-off between efficiency and accuracy when using sparse indexing features or early stopping on ActivityNet. Uniform denotes the \textsc{Uniform} baseline in Table~\ref{Table:video_level_comparison}.
    \vspace{-0.15in}}
  \vspace{-0.1in}
\end{figure}

\vspace{-0.05in}
\paragraph{Efficiency \& accuracy trade-off.} 
Before showing the results, we introduce how we use feature interpolation to further enhance the efficiency of our system. Apart from using features from all $N$ time stamps as described in Sec.~\ref{Sec:video_level_app}, we experiment with using \emph{sparse} indexing features extracted from a subset of image-audio pairs, \ie, subsampling along the time axis. Motivated by the locally-smooth action feature space~\cite{Dwibedi_2019_CVPR} and based on our empirical observation that neighboring video features can be linearly approximated well, we synthesize the missing image and audio features by computationally cheap linear interpolation to generate the full feature sequences of length $N$. Figure~\ref{fig:interpolation} shows the recognition results when using different subsampling factors. We can see that recognition remains robust to even aggressive subsampling of the indexing features.

Next we investigate early stopping as an additional means to reduce the computational cost. Instead of repeating the skimming procedure for 10 times as in the training stage, we can choose to stop early after a few recurrent steps. Figure~\ref{fig:early_stop} shows the results when stopping at different time steps. We can see that the first three steps yield sufficient cues for recognition. This suggests that we can stop around the third step with negligible accuracy loss. See Supp. for a similar observation on Mini-Sports1M.

\paragraph{Results.} We compare our approach to the following baselines and several existing methods~\cite{yeung2016end,fan2018watching,wu2019adaframe,wu2019multiagent,korbar2019scsampler}:
\vspace{-0.05in}
\begin{itemize}[leftmargin=4mm]
\itemsep0em
\item \textsc{Random:} We randomly sample 10 out of the $N$ time stamps, and average the predictions of the image-audio pairs from these selected time stamps using the distilled image-audio network.
\vspace{-0.05in}
\item \textsc{Uniform:} The same as the previous baseline except that we perform uniform sampling.
\vspace{-0.05in}
\item \textsc{Front / Center / End:} The same as before except that the first / center / last 10 time stamps are used.
\vspace{-0.05in}
\item \textsc{Dense:} We average the prediction scores from all $N$ image-audio pairs as the video-level prediction.
\vspace{-0.05in}
\item \textsc{SCSampler~\cite{korbar2019scsampler}:} We use the idea of \cite{korbar2019scsampler} and select the 10 image-audio pairs that yield the largest confidence scores from the image-audio classifier. We average their predictions to produce the video-level prediction. 
\vspace{-0.05in}
\item \textsc{LSTM}: This is a one-layer LSTM as in our model but it is trained and tested using all $N$ image-audio features as input sequentially to predict the action label from the hidden state of the last time step.
\vspace{-0.05in}
\item \textsc{Non-Recurrent}: The same as our method except that we only use a single query operation without the recurrent iterations. We directly obtain the 10 time stamps from the indexes of the 10 largest attention weights.
\vspace{-0.05in}
\end{itemize}

Table~\ref{Table:video_level_comparison} shows the results. Our method outperforms all the baselines. The low accuracy of \textsc{Random} / \textsc{Uniform} / \textsc{Front} / \textsc{Center} / \textsc{End} indicates the importance of the context-aware selection of useful moments for action recognition. Using sparse indexing features (with a subsampling factor of 5), our method outperforms \textsc{Dense} (the status quo of how most current methods obtain video-level predictions) by a large margin using only about 1/5 of its computation cost. Our method is also better and faster than \textsc{SCSampler}~\cite{korbar2019scsampler}, despite their advantage of densely evaluating prediction results on all clips. \textsc{LSTM} performs comparably to \textsc{Random}. We suspect that it fails to aggregate the information of all time stamps when the video gets very long. \textsc{Non-Recurrent} is an ablated version of our method, and it shows that the design of recursive prediction of the ``next'' interesting moment in our method is essential. Both \textsc{LSTM} and \textsc{Non-Recurrent} support our contribution as a whole framework, \ie, iterative attention based selection.

\begin{figure}
    \center
    \includegraphics[scale=0.3]{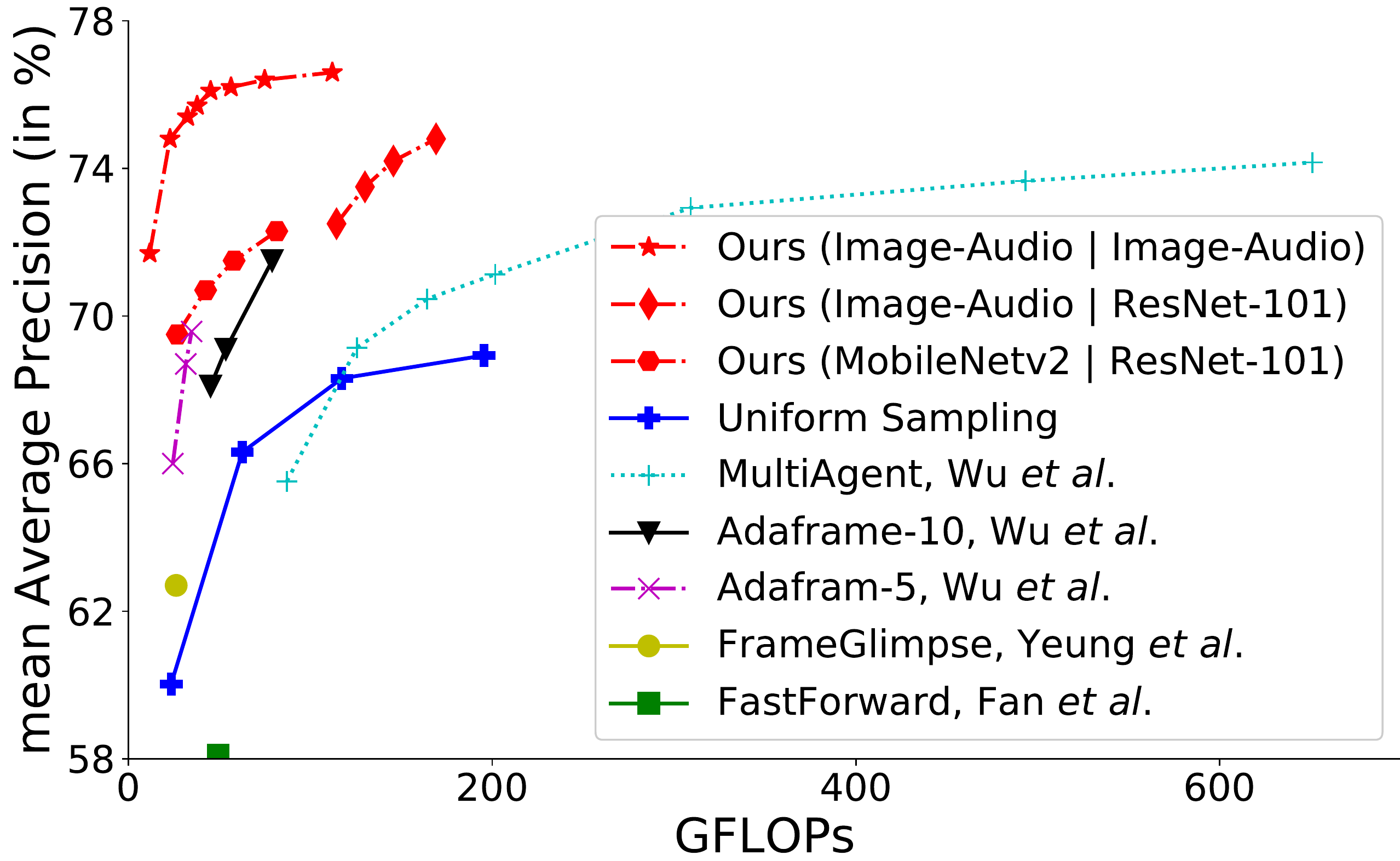}
    \caption{Comparisons with other frame selection methods on ActivityNet. We directly quote the numbers reported in AdaFrame~\cite{wu2019adaframe} and MultiAgent~\cite{wu2019multiagent} for all the baseline methods and compare the mAP against the average GFLOPs per test video. See text for details.
      \vspace{-0.15in}}
    \label{Figure:frame_selection}
    \vspace{-0.1in}
\end{figure}

\vspace{-0.05in}
\paragraph{Comparison to state of the art frame selection methods.} Fig.~\ref{Figure:frame_selection} compares our approach to state-of-the-art frame selection methods given the same computational budget. The results of the baselines are quoted from AdaFrame~\cite{wu2019adaframe} and MultiAgent~\cite{wu2019multiagent}, where they both evaluate on ActivityNet. For fair comparison, we test a variant of our method with only the visual modality, and we use the same ResNet-101 features for recognition. Our framework also has the flexibility of using cheaper features for indexing (frame selection). See Supp. for details about the single-modality architecture of our \textsc{ImgAud-Skimming} network and how we use different features for indexing and recognition. We use three different combinations denoted as {\em Ours (``indexing features'' $\vert$ ``recognition features'')} in Fig.~\ref{Figure:frame_selection}, including using MobileNetv2~\cite{sandler2018mobilenetv2} features for efficient indexing similar to~\cite{wu2019adaframe}. Moreover, to gauge the impact of our \textsc{ImgAud2Vid} step, we also report the results obtained by using image-audio features for recognition.

Our method consistently outperforms all existing methods and achieves the best balance between speed and accuracy when using the same recognition features, suggesting the accuracy boost can be attributed to our novel differentiable indexing mechanism. Furthermore, with the aid of \textsc{ImgAud2Vid} distillation, we achieve much higher accuracy with much less computation cost; this scheme combines the efficiency of our image-audio clip-level recognition with the speedup and accuracy enabled by our \textsc{ImgAud-Skimming} network for video-level recognition.

\begin{figure}
    \center
    \includegraphics[width=1\linewidth]{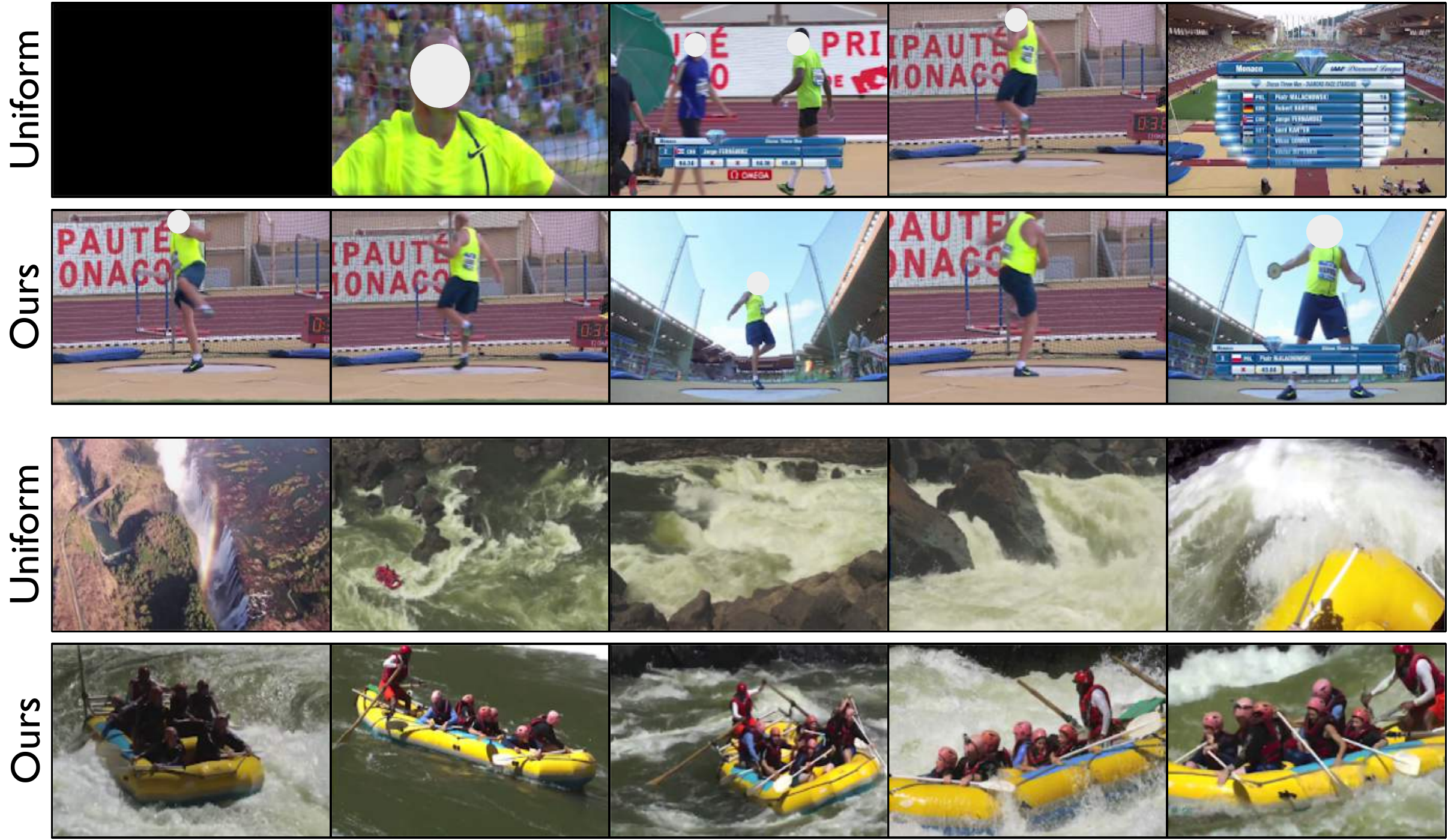}
    \caption{Qualitative examples of 5 uniformly selected moments and the first 5 visually useful moments selected by our method for two videos of the actions \emph{throwing discus} and \emph{rafting} in ActivityNet. The frames selected by our method are more indicative of the corresponding action. %See Supp.~video\textsuperscript{\ref{project_page}} for examples of acoustically useful moments selected by our method.
    \vspace{-0.2in}
    }
    \label{fig:qualitative_examples}
    \vspace{-0.1in}
\end{figure}

\paragraph{Comparison to the state of the art
on ActivityNet.} 
Having compared our skimming approach to existing methods for frame selection, now we compare to state-of-the-art activity recognition models that forgo frame selection. For fair comparison, we use the ResNet-152 model provided by~\cite{wu2019multiagent}. This model is pre-trained on ImageNet and fine-tuned on ActivityNet with TSN-style~\cite{wang2016temporal} training. As shown in Table~\ref{Table:activitynet_sota}\textcolor{blue}{a}, our method consistently outperforms all the previous state-of-the-art methods. To show that the benefits of our method extend even to more powerful but expensive features, we use R(2+1)D-152 features for recognition in Table~\ref{Table:activitynet_sota}\textcolor{blue}{b}. When using R(2+1)D-152 features for both indexing and recognition, we outperform the dense approach while being \emph{$10\times$ faster}. We can still achieve comparable performance to the dense approach if using our image-audio features for indexing, while being \emph{20$\times$ faster}.

\begin{table}
\begin{subtable}{\linewidth}\centering
{\resizebox{1\linewidth}{!}{
\begin{tabular}{zccccc}
\toprule
     & Backbone & Pre-trained & Accuracy & mAP
     \\ \hline
IDT~\cite{wang2013action}  & --  & ImageNet &  64.7 & 68.7 \\ 
C3D~\cite{tran2015learning}  & --    & Sports1M    &   65.8 & 67.7 \\ 
P3D~\cite{qiu2017learning}  &  ResNet-152  &  ImageNet &  75.1 & 78.9   \\ 
RRA~\cite{zhu2018fine}  &  ResNet-152  &  ImageNet    &  78.8 & 83.4 \\ 
MARL~\cite{wu2019multiagent} &  ResNet-152  & ImageNet &   79.8 & 83.8 \\ 
Ours &    ResNet-152      &     ImageNet  & \textbf{80.3}  &  \textbf{84.2}   \\
\bottomrule
\end{tabular}
}}
\vspace{-0.05in}
\caption{Comparison to prior work with ResNet-152 features.}
\vspace{0.05in}
\end{subtable}
\begin{subtable}{\linewidth}\centering
{\resizebox{1\linewidth}{!}{
\begin{tabular}{zccccc}
\toprule
     & Indexing & Recognition & mAP & TFLOPs
     \\ \hline
Dense  &   --  & R(2+1)D-152 &  88.9 &  25.9 \\
Uniform  &   --  & R(2+1)D-152 &  87.2 & 1.26 \\
Ours  &   Image-Audio  & R(2+1)D-152 & 88.5 & 1.31 \\
Ours  &   R(2+1)D-152 & R(2+1)D-152 & \textbf{89.9} & 2.64 \\
\bottomrule
\end{tabular}
}}
\vspace{-0.05in}
\caption{Accuracy vs. Efficiency with R(2+1)D-152 features.}
\end{subtable}
\vspace{0.05in}
\caption{ActivityNet comparison to SOTA methods.
\vspace{-0.2in}
}
\label{Table:activitynet_sota}
\vspace{-0.1in}
\end{table}

\vspace{-0.05in}
\subsection{Qualitative Analysis}
Figure~\ref{fig:qualitative_examples} shows frames selected by our method using the visual modality versus those obtained by uniform sampling. The frames chosen by our method are much more informative of the action in the video compared to those  uniformly sampled. See Supp.~video\footnote{\scriptsize\label{project_page}\url{http://vision.cs.utexas.edu/projects/listen_to_look/}} for examples of acoustically useful moments selected by our method.

We can inspect per-class performance to understand what are the classes that benefit the most from our skimming mechanism compared to uniform sampling. The top classes in descending order of accuracy gain are: cleaning sink, beer pong, gargling mouthwash, painting furniture, archery, laying tile, and triple jump---classes where the action is sporadic and is often exhibited over a short segment of the video. See Supp.~for more analysis.
%===========================================================
%Conclusion
\vspace{-0.1in}
\section{Conclusion}
We presented an approach to achieve both accurate and efficient action recognition in long untrimmed videos by leveraging audio as a previewing tool. Our \textsc{ImgAud2Vid} distillation framework replaces the expensive clip-based model by a lightweight image-audio based model, enabling efficient clip-level action recognition. Moreover, we propose an \textsc{ImgAud-Skimming} network that iteratively selects useful image-audio pairs, enabling efficient video-level action recognition. Our work strikes a favorable balance between speed and accuracy, and we achieve state-of-the-art results for video action recognition using few selected frames or clips. In future work, we plan to investigate salient spatial region selection along with our temporal frame selection, which can potentially lead to finer granularity of action understanding with improved efficiency, as well as extensions to allow the multi-label setting.
%===========================================================

%\vspace*{-0.1in}
%\footnotesize
\paragraph{Acknowledgements:} Thanks to Bruno Korbar, Zuxuan Wu, and Wenhao Wu for help with experiments and to Weiyao Wang, Du Tran, and the UT Austin vision group for helpful discussions. 

{\small
\bibliographystyle{ieee}
\bibliography{ref_RG}

\begin{thebibliography}{10}\itemsep=-1pt

\bibitem{cisco_study}
Cisco visual networking index: Forecast and trends, 2017–2022 white paper.

\bibitem{afouras2018conversation}
T.~Afouras, J.~S. Chung, and A.~Zisserman.
\newblock The conversation: Deep audio-visual speech enhancement.
\newblock In {\em Interspeech}, 2018.

\bibitem{albanie2018emotion}
S.~Albanie, A.~Nagrani, A.~Vedaldi, and A.~Zisserman.
\newblock Emotion recognition in speech using cross-modal transfer in the wild.
\newblock In {\em ACM Multimedia}, 2018.

\bibitem{alwassel2018action}
H.~Alwassel, F.~Caba~Heilbron, and B.~Ghanem.
\newblock Action search: Spotting actions in videos and its application to
  temporal action localization.
\newblock In {\em ECCV}, 2018.

\bibitem{arandjelovic2017look}
R.~Arandjelovic and A.~Zisserman.
\newblock Look, listen and learn.
\newblock In {\em ICCV}, 2017.

\bibitem{arandjelovic2017objects}
R.~Arandjelovi{\'c} and A.~Zisserman.
\newblock Objects that sound.
\newblock In {\em ECCV}, 2018.

\bibitem{aytar2016soundnet}
Y.~Aytar, C.~Vondrick, and A.~Torralba.
\newblock Soundnet: Learning sound representations from unlabeled video.
\newblock In {\em NeurIPS}, 2016.

\bibitem{buch2017sst}
S.~Buch, V.~Escorcia, C.~Shen, B.~Ghanem, and J.~Carlos~Niebles.
\newblock Sst: Single-stream temporal action proposals.
\newblock In {\em CVPR}, 2017.

\bibitem{caba2015activitynet}
F.~Caba~Heilbron, V.~Escorcia, B.~Ghanem, and J.~Carlos~Niebles.
\newblock Activitynet: A large-scale video benchmark for human activity
  understanding.
\newblock In {\em CVPR}, 2015.

\bibitem{carreira2017quo}
J.~Carreira and A.~Zisserman.
\newblock Quo vadis, action recognition? a new model and the kinetics dataset.
\newblock In {\em CVPR}, 2017.

\bibitem{chen2018multi}
Y.~Chen, Y.~Kalantidis, J.~Li, S.~Yan, and J.~Feng.
\newblock Multi-fiber networks for video recognition.
\newblock In {\em ECCV}, 2018.

\bibitem{donahue2015long}
J.~Donahue, L.~Anne~Hendricks, S.~Guadarrama, M.~Rohrbach, S.~Venugopalan,
  K.~Saenko, and T.~Darrell.
\newblock Long-term recurrent convolutional networks for visual recognition and
  description.
\newblock In {\em CVPR}, 2015.

\bibitem{Dwibedi_2019_CVPR}
D.~Dwibedi, Y.~Aytar, J.~Tompson, P.~Sermanet, and A.~Zisserman.
\newblock Temporal cycle-consistency learning.
\newblock In {\em CVPR}, 2019.

\bibitem{ephrat2018looking}
A.~Ephrat, I.~Mosseri, O.~Lang, T.~Dekel, K.~Wilson, A.~Hassidim, W.~T.
  Freeman, and M.~Rubinstein.
\newblock Looking to listen at the cocktail party: A speaker-independent
  audio-visual model for speech separation.
\newblock In {\em SIGGRAPH}, 2018.

\bibitem{fan2018watching}
H.~Fan, Z.~Xu, L.~Zhu, C.~Yan, J.~Ge, and Y.~Yang.
\newblock Watching a small portion could be as good as watching all: Towards
  efficient video classification.
\newblock In {\em IJCAI}, 2018.

\bibitem{feichtenhofer2019slowfast}
C.~Feichtenhofer, H.~Fan, J.~Malik, and K.~He.
\newblock Slowfast networks for video recognition.
\newblock In {\em ICCV}, 2019.

\bibitem{feichtenhofer2016convolutional}
C.~Feichtenhofer, A.~Pinz, and A.~Zisserman.
\newblock Convolutional two-stream network fusion for video action recognition.
\newblock In {\em CVPR}, 2016.

\bibitem{fernando2015modeling}
B.~Fernando, E.~Gavves, J.~M. Oramas, A.~Ghodrati, and T.~Tuytelaars.
\newblock Modeling video evolution for action recognition.
\newblock In {\em CVPR}, 2015.

\bibitem{gan2019tracking}
C.~Gan, H.~Zhao, P.~Chen, D.~Cox, and A.~Torralba.
\newblock Self-supervised moving vehicle tracking with stereo sound.
\newblock In {\em ICCV}, 2019.

\bibitem{gao2018objectSounds}
R.~Gao, R.~Feris, and K.~Grauman.
\newblock Learning to separate object sounds by watching unlabeled video.
\newblock In {\em ECCV}, 2018.

\bibitem{gao2019visualsound}
R.~Gao and K.~Grauman.
\newblock 2.5d visual sound.
\newblock In {\em CVPR}, 2019.

\bibitem{gao2019coseparation}
R.~Gao and K.~Grauman.
\newblock Co-separating sounds of visual objects.
\newblock In {\em ICCV}, 2019.

\bibitem{gaver1993world}
W.~W. Gaver.
\newblock What in the world do we hear?: An ecological approach to auditory
  event perception.
\newblock {\em Ecological psychology}, 1993.

\bibitem{Girdhar_17a_ActionVLAD}
R.~Girdhar, D.~Ramanan, A.~Gupta, J.~Sivic, and B.~Russell.
\newblock {ActionVLAD}: Learning spatio-temporal aggregation for action
  classification.
\newblock In {\em CVPR}, 2017.

\bibitem{girdhar2019distinit}
R.~Girdhar, D.~Tran, L.~Torresani, and D.~Ramanan.
\newblock Distinit: Learning video representations without a single labeled
  video.
\newblock In {\em ICCV}, 2019.

\bibitem{gong2014diverse}
B.~Gong, W.-L. Chao, K.~Grauman, and F.~Sha.
\newblock Diverse sequential subset selection for supervised video
  summarization.
\newblock In {\em NeurIPS}, 2014.

\bibitem{graves2014neural}
A.~Graves, G.~Wayne, and I.~Danihelka.
\newblock Neural turing machines.
\newblock {\em arXiv preprint arXiv:1410.5401}, 2014.

\bibitem{gupta2016cross}
S.~Gupta, J.~Hoffman, and J.~Malik.
\newblock Cross modal distillation for supervision transfer.
\newblock In {\em CVPR}, 2016.

\bibitem{hinton2015distilling}
G.~Hinton, O.~Vinyals, and J.~Dean.
\newblock Distilling the knowledge in a neural network.
\newblock {\em arXiv preprint arXiv:1503.02531}, 2015.

\bibitem{jain2013representing}
A.~Jain, A.~Gupta, M.~Rodriguez, and L.~S. Davis.
\newblock Representing videos using mid-level discriminative patches.
\newblock In {\em CVPR}, 2013.

\bibitem{jain2014action}
M.~Jain, J.~Van~Gemert, H.~J{\'e}gou, P.~Bouthemy, and C.~G. Snoek.
\newblock Action localization with tubelets from motion.
\newblock In {\em CVPR}, 2014.

\bibitem{karpathy2014large}
A.~Karpathy, G.~Toderici, S.~Shetty, T.~Leung, R.~Sukthankar, and L.~Fei-Fei.
\newblock Large-scale video classification with convolutional neural networks.
\newblock In {\em CVPR}, 2014.

\bibitem{kay2017kinetics}
W.~Kay, J.~Carreira, K.~Simonyan, B.~Zhang, C.~Hillier, S.~Vijayanarasimhan,
  F.~Viola, T.~Green, T.~Back, P.~Natsev, et~al.
\newblock The kinetics human action video dataset.
\newblock {\em arXiv preprint arXiv:1705.06950}, 2017.

\bibitem{kazakos2019TBN}
E.~Kazakos, A.~Nagrani, A.~Zisserman, and D.~Damen.
\newblock Epic-fusion: Audio-visual temporal binding for egocentric action
  recognition.
\newblock In {\em ICCV}, 2019.

\bibitem{Korbar2018cotraining}
B.~Korbar, D.~Tran, and L.~Torresani.
\newblock Co-training of audio and video representations from self-supervised
  temporal synchronization.
\newblock In {\em NeurIPS}, 2018.

\bibitem{korbar2019scsampler}
B.~Korbar, D.~Tran, and L.~Torresani.
\newblock Scsampler: Sampling salient clips from video for efficient action
  recognition.
\newblock In {\em ICCV}, 2019.

\bibitem{laptev2003space}
I.~Laptev and T.~Lindeberg.
\newblock Space-time interest points.
\newblock In {\em ICCV}, 2003.

\bibitem{lee2012discovering}
Y.~J. Lee, J.~Ghosh, and K.~Grauman.
\newblock Discovering important people and objects for egocentric video
  summarization.
\newblock In {\em CVPR}, 2012.

\bibitem{lin2019tsm}
J.~Lin, C.~Gan, and S.~Han.
\newblock Tsm: Temporal shift module for efficient video understanding.
\newblock In {\em ICCV}, 2019.

\bibitem{lin2018bsn}
T.~Lin, X.~Zhao, H.~Su, C.~Wang, and M.~Yang.
\newblock Bsn: Boundary sensitive network for temporal action proposal
  generation.
\newblock In {\em ECCV}, 2018.

\bibitem{long2018attention}
X.~Long, C.~Gan, G.~De~Melo, J.~Wu, X.~Liu, and S.~Wen.
\newblock Attention clusters: Purely attention based local feature integration
  for video classification.
\newblock In {\em CVPR}, 2018.

\bibitem{mahasseni2017unsupervised}
B.~Mahasseni, M.~Lam, and S.~Todorovic.
\newblock Unsupervised video summarization with adversarial lstm networks.
\newblock In {\em CVPR}, 2017.

\bibitem{mnih2014recurrent}
V.~Mnih, N.~Heess, A.~Graves, et~al.
\newblock Recurrent models of visual attention.
\newblock In {\em NeurIPS}, 2014.

\bibitem{morgadoNIPS18}
P.~Morgado, N.~Vasconcelos, T.~Langlois, and O.~Wang.
\newblock Self-supervised generation of spatial audio for 360${}^\circ$ video.
\newblock In {\em NeurIPS}, 2018.

\bibitem{owens2018audio}
A.~Owens and A.~A. Efros.
\newblock Audio-visual scene analysis with self-supervised multisensory
  features.
\newblock In {\em ECCV}, 2018.

\bibitem{owens2016visually}
A.~Owens, P.~Isola, J.~McDermott, A.~Torralba, E.~H. Adelson, and W.~T.
  Freeman.
\newblock Visually indicated sounds.
\newblock In {\em CVPR}, 2016.

\bibitem{owens2016ambient}
A.~Owens, J.~Wu, J.~H. McDermott, W.~T. Freeman, and A.~Torralba.
\newblock Ambient sound provides supervision for visual learning.
\newblock In {\em ECCV}, 2016.

\bibitem{pirsiavash2014parsing}
H.~Pirsiavash and D.~Ramanan.
\newblock Parsing videos of actions with segmental grammars.
\newblock In {\em CVPR}, 2014.

\bibitem{qiu2017learning}
Z.~Qiu, T.~Yao, and T.~Mei.
\newblock Learning spatio-temporal representation with pseudo-3d residual
  networks.
\newblock In {\em ICCV}, 2017.

\bibitem{raptis2012discovering}
M.~Raptis, I.~Kokkinos, and S.~Soatto.
\newblock Discovering discriminative action parts from mid-level video
  representations.
\newblock In {\em CVPR}, 2012.

\bibitem{sandler2018mobilenetv2}
M.~Sandler, A.~Howard, M.~Zhu, A.~Zhmoginov, and L.-C. Chen.
\newblock Mobilenetv2: Inverted residuals and linear bottlenecks.
\newblock In {\em CVPR}, 2018.

\bibitem{Senocak_2019_PAMI}
A.~{Senocak}, T.-H. {Oh}, J.~{Kim}, M.~{Yang}, and I.~S. {Kweon}.
\newblock Learning to localize sound sources in visual scenes: Analysis and
  applications.
\newblock {\em TPAMI}, 2019.

\bibitem{shou2017cdc}
Z.~Shou, J.~Chan, A.~Zareian, K.~Miyazawa, and S.-F. Chang.
\newblock Cdc: Convolutional-de-convolutional networks for precise temporal
  action localization in untrimmed videos.
\newblock In {\em CVPR}, 2017.

\bibitem{shou2019dmc}
Z.~Shou, X.~Lin, Y.~Kalantidis, L.~Sevilla-Lara, M.~Rohrbach, S.-F. Chang, and
  Z.~Yan.
\newblock Dmc-net: Generating discriminative motion cues for fast compressed
  video action recognition.
\newblock In {\em CVPR}, 2019.

\bibitem{twostream}
K.~Simonyan and A.~Zisserman.
\newblock Two-stream convolutional networks for action recognition in videos.
\newblock In {\em NeurIPS}, 2014.

\bibitem{soomro2012ucf101}
K.~Soomro, A.~R. Zamir, and M.~Shah.
\newblock Ucf101: A dataset of 101 human actions classes from videos in the
  wild.
\newblock {\em arXiv preprint arXiv:1212.0402}, 2012.

\bibitem{su2016leaving}
Y.-C. Su and K.~Grauman.
\newblock Leaving some stones unturned: dynamic feature prioritization for
  activity detection in streaming video.
\newblock In {\em ECCV}, 2016.

\bibitem{sukhbaatar2015end}
S.~Sukhbaatar, J.~Weston, R.~Fergus, et~al.
\newblock End-to-end memory networks.
\newblock In {\em NeurIPS}, 2015.

\bibitem{sun2019videobert}
C.~Sun, A.~Myers, C.~Vondrick, K.~Murphy, and C.~Schmid.
\newblock Videobert: A joint model for video and language representation
  learning.
\newblock In {\em ICCV}, 2019.

\bibitem{tian2018audio}
Y.~Tian, J.~Shi, B.~Li, Z.~Duan, and C.~Xu.
\newblock Audio-visual event localization in unconstrained videos.
\newblock In {\em ECCV}, 2018.

\bibitem{tran2015learning}
D.~Tran, L.~Bourdev, R.~Fergus, L.~Torresani, and M.~Paluri.
\newblock Learning spatiotemporal features with 3d convolutional networks.
\newblock In {\em ICCV}, 2015.

\bibitem{tran2018closer}
D.~Tran, H.~Wang, L.~Torresani, J.~Ray, Y.~LeCun, and M.~Paluri.
\newblock A closer look at spatiotemporal convolutions for action recognition.
\newblock In {\em CVPR}, 2018.

\bibitem{varol2016long}
G.~Varol, I.~Laptev, and C.~Schmid.
\newblock Long-term temporal convolutions for action recognition.
\newblock {\em TPAMI}, 2017.

\bibitem{vaswani2017attention}
A.~Vaswani, N.~Shazeer, N.~Parmar, J.~Uszkoreit, L.~Jones, A.~N. Gomez,
  {\L}.~Kaiser, and I.~Polosukhin.
\newblock Attention is all you need.
\newblock In {\em NeurIPS}, 2017.

\bibitem{vinyals2015pointer}
O.~Vinyals, M.~Fortunato, and N.~Jaitly.
\newblock Pointer networks.
\newblock In {\em NeurIPS}, 2015.

\bibitem{wang2013action}
H.~Wang and C.~Schmid.
\newblock Action recognition with improved trajectories.
\newblock In {\em CVPR}, 2013.

\bibitem{wang2013motionlets}
L.~Wang, Y.~Qiao, and X.~Tang.
\newblock Motionlets: Mid-level 3d parts for human motion recognition.
\newblock In {\em CVPR}, 2013.

\bibitem{wang2017untrimmednets}
L.~Wang, Y.~Xiong, D.~Lin, and L.~Van~Gool.
\newblock Untrimmednets for weakly supervised action recognition and detection.
\newblock In {\em CVPR}, 2017.

\bibitem{wang2016temporal}
L.~Wang, Y.~Xiong, Z.~Wang, Y.~Qiao, D.~Lin, X.~Tang, and L.~Van~Gool.
\newblock Temporal segment networks: Towards good practices for deep action
  recognition.
\newblock In {\em ECCV}, 2016.

\bibitem{wang2019makes}
W.~Wang, D.~Tran, and M.~Feiszli.
\newblock What makes training multi-modal networks hard?
\newblock {\em arXiv preprint arXiv:1905.12681}, 2019.

\bibitem{wang2018non}
X.~Wang, R.~Girshick, A.~Gupta, and K.~He.
\newblock Non-local neural networks.
\newblock In {\em CVPR}, 2018.

\bibitem{willems2008efficient}
G.~Willems, T.~Tuytelaars, and L.~Van~Gool.
\newblock An efficient dense and scale-invariant spatio-temporal interest point
  detector.
\newblock In {\em ECCV}, 2008.

\bibitem{wu2019long}
C.-Y. Wu, C.~Feichtenhofer, H.~Fan, K.~He, P.~Krahenbuhl, and R.~Girshick.
\newblock Long-term feature banks for detailed video understanding.
\newblock In {\em CVPR}, 2019.

\bibitem{wu2018compressed}
C.-Y. Wu, M.~Zaheer, H.~Hu, R.~Manmatha, A.~J. Smola, and
  P.~Kr{\"a}henb{\"u}hl.
\newblock Compressed video action recognition.
\newblock In {\em CVPR}, 2018.

\bibitem{wu2019multiagent}
W.~Wu, D.~He, X.~Tan, S.~Chen, and S.~Wen.
\newblock Multi-agent reinforcement learning based frame sampling for effective
  untrimmed video recognition.
\newblock In {\em ICCV}, 2019.

\bibitem{wu2016multi}
Z.~Wu, Y.-G. Jiang, X.~Wang, H.~Ye, and X.~Xue.
\newblock Multi-stream multi-class fusion of deep networks for video
  classification.
\newblock In {\em ACM-MM}, 2016.

\bibitem{wu2019adaframe}
Z.~Wu, C.~Xiong, C.-Y. Ma, R.~Socher, and L.~S. Davis.
\newblock Adaframe: Adaptive frame selection for fast video recognition.
\newblock In {\em CVPR}, 2019.

\bibitem{xie2018rethinking}
S.~Xie, C.~Sun, J.~Huang, Z.~Tu, and K.~Murphy.
\newblock Rethinking spatiotemporal feature learning: Speed-accuracy trade-offs
  in video classification.
\newblock In {\em ECCV}, 2018.

\bibitem{xu2017r}
H.~Xu, A.~Das, and K.~Saenko.
\newblock R-c3d: Region convolutional 3d network for temporal activity
  detection.
\newblock In {\em ICCV}, 2017.

\bibitem{yeung2016end}
S.~Yeung, O.~Russakovsky, G.~Mori, and L.~Fei-Fei.
\newblock End-to-end learning of action detection from frame glimpses in
  videos.
\newblock In {\em CVPR}, 2016.

\bibitem{yue2015beyond}
J.~Yue-Hei~Ng, M.~Hausknecht, S.~Vijayanarasimhan, O.~Vinyals, R.~Monga, and
  G.~Toderici.
\newblock Beyond short snippets: Deep networks for video classification.
\newblock In {\em CVPR}, 2015.

\bibitem{zhang2018retrospective}
K.~Zhang, K.~Grauman, and F.~Sha.
\newblock Retrospective encoders for video summarization.
\newblock In {\em ECCV}, 2018.

\bibitem{zhao2018sound}
H.~Zhao, C.~Gan, A.~Rouditchenko, C.~Vondrick, J.~McDermott, and A.~Torralba.
\newblock The sound of pixels.
\newblock In {\em ECCV}, 2018.

\bibitem{zhao2017temporal}
Y.~Zhao, Y.~Xiong, L.~Wang, Z.~Wu, X.~Tang, and D.~Lin.
\newblock Temporal action detection with structured segment networks.
\newblock In {\em ICCV}, 2017.

\bibitem{zhou2018temporal}
B.~Zhou, A.~Andonian, A.~Oliva, and A.~Torralba.
\newblock Temporal relational reasoning in videos.
\newblock In {\em ECCV}, 2018.

\bibitem{Zhou_2019_ICCV}
H.~Zhou, Z.~Liu, X.~Xu, P.~Luo, and X.~Wang.
\newblock Vision-infused deep audio inpainting.
\newblock In {\em ICCV}, 2019.

\bibitem{zhou2017visual}
Y.~Zhou, Z.~Wang, C.~Fang, T.~Bui, and T.~L. Berg.
\newblock Visual to sound: Generating natural sound for videos in the wild.
\newblock In {\em CVPR}, 2018.

\bibitem{zhu2018fine}
C.~Zhu, X.~Tan, F.~Zhou, X.~Liu, K.~Yue, E.~Ding, and Y.~Ma.
\newblock Fine-grained video categorization with redundancy reduction
  attention.
\newblock In {\em ECCV}, 2018.

\bibitem{zolfaghari2018eco}
M.~Zolfaghari, K.~Singh, and T.~Brox.
\newblock Eco: Efficient convolutional network for online video understanding.
\newblock In {\em ECCV}, 2018.

\end{thebibliography}
}

\end{document}